\documentclass[11pt]{article}
\usepackage{coling2018}
\usepackage{times}
\usepackage{url}
\usepackage{latexsym}

\usepackage[utf8]{inputenc}
\usepackage[T1]{fontenc}
\usepackage{multirow}

\usepackage{amsmath}
\usepackage{colortbl}
\usepackage{hhline}

\usepackage{enumitem}

\makeatletter
\newcommand{\@BIBLABEL}{\@emptybiblabel}
\newcommand{\@emptybiblabel}[1]{}
\makeatother
\usepackage{hyperref}

\newcommand{\tabincell}[2]{\begin{tabular}{@{}#1@{}}#2\end{tabular}}

\definecolor{UUlight}{RGB}{200,200,200}
\definecolor{UUmedium}{RGB}{90,90,90}
\definecolor{UUdark}{RGB}{80,80,80}
\definecolor{UUred}{RGB}{153,0,0}

\title{An Evaluation of Neural Machine Translation Models \\on Historical Spelling Normalization}

\author{Gongbo Tang, Fabienne Cap, Eva Pettersson, and Joakim Nivre \\
  Uppsala University  \\Uppsala, Sweden\\
  {\tt firstname.lastname@lingfil.uu.se} \\
  }

\date{}

\begin{document}
\maketitle
\begin{abstract}
  In this paper, we apply different NMT models to the problem of historical spelling normalization for five languages: English, German, Hungarian, Icelandic, and Swedish. The NMT models are at different levels, have different attention mechanisms, and different neural network architectures. Our results show that NMT models are much better than SMT models in terms of character error rate. The vanilla RNNs are competitive to GRUs/LSTMs in historical spelling normalization. Transformer models perform better only when provided with more training data. We also find that subword-level models with a small subword vocabulary are better than character-level models for low-resource languages. In addition, we propose a hybrid method which further improves the performance of historical spelling normalization. 
\end{abstract}

\section{Introduction}

\blfootnote{This work is licensed under a Creative Commons Attribution 4.0 International License. License
details: \url{http://creativecommons.org/licenses/by/4.0/}}

With increasing access to digital historical text, the processing of these 
historical texts is attracting more and more interest. 
However, in contrast to modern text, historical text processing faces 
more challenges. First, for historical text, there is little annotated data 
for training a model, which leads to data sparsity issues when using statistical 
methods, similar to the situation for low-resource languages. 
Second, there are a lot of variations in historical texts from different time 
periods, not only in spelling but also in lexical semantics and syntax.
Therefore, the NLP tools developed for modern 
text cannot be used for these historical texts directly.
Spelling normalization is the task of mapping a historical spelling 
to its modern spelling. It is usually used as a preprocessing step
before feeding the historical text into modern NLP tools 
\cite{pettersson2013smt,bollmann2013pos,sanchez2013open}, 
which leads to much better results 
compared to analyzing unnormalized historical texts. 

There are some papers in which neural machine translation (NMT) models are employed for the spelling 
normalization task. \newcite{korchagina2017normalizing} utilizes a 
character-level NMT model for medieval German texts. 
\newcite{bollmann2017learning} apply an attention-based NMT model to 
historical German texts.
The evidence so far is too incomplete to draw any general conclusions about 
the utility of different NMT models for historical spelling normalization. 
We are interested in exploring how different properties of NMT models interact with 
different aspects of the spelling normalization problem and find some generalizations 
about the use of NMT models for this task.

In this paper, we apply different NMT models to the spelling normalization task 
for historical stages of five languages, English, German, Hungarian, Icelandic, and Swedish. 
We compare our result to those of \newcite{pettersson2014evaluation}, 
which are obtained with statistical machine translation (SMT) models. 
We investigate whether NMT models outperform SMT models in general, and 
explore which properties of NMT models are suitable for spelling normalization.
Compared to the conventional machine translation (MT) tasks, 
we train models on token pairs instead of sentence pairs. 
Token length is usually shorter than sentence length. After reviewing related work in Section 2, 
we give our hypotheses about utilizing NMT models for the spelling 
normalization task and select different NMT models based on our hypotheses in Section 3.
The selected NMT models are at different levels (character-level, subword-level), 
have different attention mechanisms (no attention, soft-attention, multi-head-attention), 
and different neural network architectures (vanilla recurrent neural networks 
(RNNs), gated recurrent units (GRUs), long short-term memory units (LSTMs), and 
self-attention).
In Section 4, we describe the datasets and our detailed experimental settings. 
In Section 5, we give our results and analyze the performance of different NMT 
models. Our conclusions and future work are in Section 6.

To conclude, our main contributions can be summarized as follows:
\begin{itemize}[noitemsep]
  \item We evaluate different NMT models on historical spelling normalization in a multilingual setting. 
  \item We find that NMT models are better than SMT models considering character error rate (CER). 
  \item We show that vanilla RNNs are competitive to GRUs/LSTMs. 
  \item We demonstrate that Transformer models perform better when provided with more training data. 
  \item We reveal that models with a small subword vocabulary are better than character-level models for low-resource languages. 
\end{itemize}

\section{Related Work}
\subsection {Historical Spelling Normalization}
Various methods have been employed for historical spelling normalization. 
\newcite{rayson2005vard} use a dictionary to map tokens to their modernized spellings,
and many different edit-distance-based methods have been proposed to deal with spelling 
normalization \cite{bollmann2011rule,pettersson2013normalisation}.
In addition, character-level SMT models have been applied to spelling normalization, where models are trained on token pairs instead of sentence pairs
\cite{pettersson2013smt,scherrer2013modernizing,sanchez2013open}. 
Each character of a token is viewed as a word of a sentence. 
The language models are trained on character N-grams instead of word N-grams.
\newcite{pettersson2014evaluation} evaluate dictionary-based methods, 
edit distance-based methods, and SMT methods on five different historical languages. 
The results show that the character-level SMT model performs best on four 
out of five historical languages.

With the development of deep learning, various neural networks have been 
applied to many tasks. 
In recent years, NMT models 
\cite{kal2013recurrent,sutskever2014sequence,cho2014learning,bahdanau15joint} 
have outperformed SMT models \cite{koehn2003statistical} distinctly 
in various translation tasks. We hypothesize that NMT models also perform 
better than SMT models for the historical spelling normalization task.
\newcite{bollmann2016improve} view the spelling normalization as a 
character-level sequence labeling task, and utilize a bi-directional LSTM for 
this task, which is better than a conditional random field (CRF) model. 
They also use 
additional data with similar but not the same historical spelling for 
a multi-task learning model, and gain further improvement.
\newcite{korchagina2017normalizing} applies a character-level NMT model to 
medieval German text, and finds that the NMT models can only outperform the 
SMT models with a larger training set.
\newcite{bollmann2017learning} test attention-based NMT models, 
and multi-task learning models which learn to normalize and pronounce with 
a grapheme-to-phoneme dictionary, on spelling normalization. Both of them achieve 
good performance. They hypothesize that the reason why the combination of 
these two models does not gain more improvement is that the multi-task 
learning has already learned the attention patterns.

\subsection {Neural Machine Translation}
In vanilla NMT models, the source sentence is encoded into a fixed-size 
vector by the encoder. Then, this vector is fed into a decoder. The decoder 
generates the target sentence word by word conditioned on the fixed-size vector 
and the generated target words \cite{kal2013recurrent}. 
Various RNN architectures are usually used as encoders and decoders. 
\newcite{cho2014learning} find that the vanilla RNN-based NMT models 
perform poorly in translating long sentences, which means that 
vanilla RNNs have problems with long-distance dependencies. 
To deal with these problems, 
\newcite{cho2014learning} propose GRUs, while \newcite{sutskever2014sequence} use 
LSTMs \cite{hochreiter1997long} to replace the vanilla RNNs.
However, any two tokens in RNNs still have a linear distance. Thus, 
\newcite{vaswani2017Attention} replace RNNs with self-attention networks which 
connect any two tokens in a sentence directly. 

Due to the expensive computation of NMT models, the vocabulary size is 
usually very limited, which causes a lot of out-of-vocabulary (OOV) words. 
Character-level models \cite{ling15char,Costa2016Character,chung2016character} 
and subword-level models \cite{sennrich16sub,wu2016google} are widely 
used to deal with OOV problems.
These two kinds of models need additional segmentation compared to word-level models.
For character-level models, we just need to separate each character by space. 
But we need more complicated segmentation methods for subword-level models.
\newcite{sennrich16sub} utilize character n-grams and a byte pair encoding (BPE) 
algorithm \cite{gage1994new} for segmentation. 
\newcite{wu2016google} apply the wordpiece model \cite{schuster2012japanese} 
to segmentation. Based on their experiments, subword-level models 
outperform word-level and character-level models. 

Attention-based NMT models have outperformed all the other architectures 
in NMT in recent years. Many improved attention-based models have been 
proposed. 
\newcite{bahdanau15joint} propose an attention-based model which can 
automatically search for source words that their hidden states are relevant to predicting a 
target word during decoding. Source words which have a higher correlation 
with the predicting target word will be assigned a higher weight. Most 
of the attention-based models use this kind of attention, 
which is called soft-attention in \newcite{xu2015show}. 
\newcite{vaswani2017Attention} propose a model named \textit{Transformer}, with multi-layer 
and multi-head attention mechanism which is more fine-grained. 

\section{NMT Models}
When we apply NMT models to the historical spelling normalization task, 
the first research question is which NMT model is most suitable for this task. 
In this section, we first give four hypotheses about NMT models for spelling 
normalization, based on the data features of historical spellings and the 
features of NMT models. Then, we list 8 different NMT models to consider for 
the spelling normalization task. 

\subsection{Hypotheses}
\paragraph{Hypothesis 1} The performance gap between vanilla RNNs and GRUs/LSTMs is small. 
In contrast to conventional NMT models, the historical and modern token pairs 
are our training data instead of parallel sentence pairs. 
In our experiments, the average token length 
varies from 4 to 6, which means that we build the model on much shorter sequences. 
The long-distance problem will be alleviated. 
It should be noted that we compare the gap to the gap in NMT \cite{bahdanau15joint}\footnote{The BLEU \cite{papineni2002bleu} scores are 15.73, and 21.83.}.

\paragraph{Hypothesis 2} The gap between NMT models with attention and without 
attention is also small.
Since the average token length is only around five, additionally
paying attention to all the tokens in the source sentence may be unnecessary. 
Thus, we hypothesize that the decoder in the vanilla Encoder-Decoder 
model can predict most of the targets correctly with only one fixed-size vector 
from the encoder, even without any attention mechanisms.
It should be mentioned that we compare the gap to the gap in NMT \cite{britz2017massive}\footnote{The BLEU scores are 17.82, and 26.75.}.

\paragraph{Hypothesis 3} Transformer models perform better than soft-attention-based models. 
Transformer models have more advanced self-attention networks and more fine-grained 
multi-head attention mechanisms compared to RNN-based models with soft-attention. 
Thus, Transformer models have better performance in conventional translation tasks. 
We hypothesize that it is the same in the spelling normalization task. 

\paragraph{Hypothesis 4} Subword-level NMT models perform better than character-level 
NMT models. 
Character-level and subword-level models are proposed to deal with the problem of 
out-of-vocabulary words mainly, and subword-level NMT models usually 
outperform character-level models. 
As we only have small sets of token pairs, it is better to use character-level or 
subword-level NMT models rather than word-level models.

\subsection{Models}
To test our hypotheses proposed in the previous section, 
we will explore 8 different NMT models for the spelling normalization task. 
The NMT models vary in attention mechanism, neural network architecture, and token granularity.
Table~\ref{table-models} gives a more detailed overview.
\begin{table}[htbp]
\begin{center}
\begin{tabular}{|c|c|c|c|}
\hline \bf Name &\bf Level &\bf Attention &\bf Architecture \\ 
\hline NoAtt-RNN & \multirow{7}{*}{character} & \multirow{3}{*}{no}& RNN\\ 
\cline{1-1}\cline{4-4} NoAtt-GRU & \ & \ &GRU\\ 
\cline{1-1}\cline{4-4} NoAtt-LSTM & \ & \ &LSTM\\ 
\cline{1-1}\cline{3-4} Att-RNN & \ & \multirow{3}{*}{soft}& RNN\\ 
\cline{1-1}\cline{4-4} Att-GRU & \ & \ & GRU\\ 
\cline{1-1}\cline{4-4} Att-LSTM & \ & \ & LSTM \\ 
\cline{1-1}\cline{3-4} Transformer & \ & multi-head& Self-attention \\ 
\hline BPE-Soft & subword & soft& LSTM\\ 
\hline
\end{tabular}
\caption{\label{table-models} NMT models for the spelling normalization task. 
\textit{RNN} means vanilla RNNs.}
\end{center}
\end{table}

\noindent 
Since we have hypothesized that different RNN architectures have slight differences, 
all the subword-level models with soft-attention are trained on LSTMs. 

\section{Experimental Setup}
\subsection{Data}
All the datasets\footnote{\href{http://stp.lingfil.uu.se/histcorp/tools.html}{http://stp.lingfil.uu.se/histcorp/tools.html}} are exactly the same as 
the parallel datasets for the SMT models in \newcite{pettersson2014evaluation}, 
which are described in Table~\ref{table-data-src}. 
Data details are shown in Table~\ref{table-data}.
The datasets consist of a list of token pairs, which have one historical spelling 
and the corresponding modernized spelling. Note that the same modern spelling may occur with different historical 
spellings. Moreover, some historical words may be extinct, and people 
have to use a spelling with a similar meaning but different lexemes as its normalization. 
Some illustrative English examples are given in Table~\ref{table-en-example}. 
If a historical spelling is identical to its modern spellings, 
we call it an \emph{unchanged spelling}. 
Otherwise, it is called a \emph{changed spelling}. 
In different languages, the number of unchanged spellings is different. 

\begin{table*}[htbp]
\begin{center}
\begin{tabular}{|l|c|l|}
\hline \bf Language  & \bf Time period &\multicolumn{1}{c|}{\bf Origin} \\
\hline English &1386--1698&\tabincell{l}{\textit{Innsbruck Corpus of English Letters}, a subset of the \textit{Innsbruck}\\ \textit{Computer Archive if Machine-Readable English Texts} \cite{markus1999manual}}\\ 
\hline German&1659--1780&\textit{GerManC} corpus \cite{scheible2011germanc}\\
\hline Hungarian&1440--1541&\textit{Hungarian Generative Diachronic Syntax} project \cite{simon2014corpus}\\
\hline Icelandic&1150--2008&\textit{Icelandic Parsed Historical Corpus} \cite{rognvaldsson2012icelandic}\\
\hline Swedish&1527--1812&\textit{Gender and Work corpus} 
(GaW) \cite{fiebranz2011making}\\
\hline
\end{tabular}
\caption{\label{table-data-src} Origin and time periods of the datasets.}
\end{center}
\end{table*}

\begin{table}[ht]
\begin{center}
\begin{tabular}{|c|c|c|c|c|c|c|c|c|}
\hline \bf Language & \bf Training &\bf  Development &\bf  Test &\bf  Unchanged&\bf Token &\bf Char &\bf Max &\bf  Avg\\
\hline English &148,852 & 16,461 & 17,791 & 75.8 & 22,302& 102&22 &4.16 \\
\hline German &\phantom{0}39,887 &\phantom{0}5,418 & \phantom{0}5,005 & 84.4& 11,521&100 &27 &4.74 \\
\hline Hungarian & 137,669 &  17,181& 17,214& 17.1 &69,624 & 128& 27& 5.91\\
\hline Icelandic & \phantom{0}52,440 &\phantom{0}6,443 & \phantom{0}6,384& 50.5& 14,845& \phantom{0}89 &16 & 4.14\\
\hline Swedish &28,327 & \phantom{0}2,590&33,544 & 64.6& 11,129& \phantom{0}92&36 &4.55 \\
\hline
\end{tabular}
\caption{\label{table-data} Statistics of the datasets. The figures in \textit{Training}, 
\textit{Development}, and \textit{Test} are the numbers of token pairs. 
The \textit{Unchanged} (\%) means the rate of unchanged spellings in the test set. 
\textit{Token} and \textit{Char} show the token and the character 
vocabulary sizes in the training set. 
\textit{Max} and \textit{Avg} show the max length and average length 
of token in the training set. 
All counts are based on case-sensitive data.} 
\end{center}
\end{table}

\begin{table}[htbp]
\begin{center}
\begin{tabular}{|c|c|c|c|c|}
\hline \bf Historical & cit\textcolor{UUmedium}{ee} & g\textcolor{UUmedium}{y}ve & g\textcolor{UUmedium}{yf} & late\\ 
\hline \bf Modern & cit\textcolor{UUmedium}{y} & g\textcolor{UUmedium}{i}ve & g\textcolor{UUmedium}{ive} & late \\ 
\hline
\end{tabular}
\caption{\label{table-en-example} Token pair examples in English.}
\end{center}
\end{table}

\noindent 
From Table~\ref{table-data}, we can see that English and Hungarian have more training data, around 140,000 token pairs, while Swedish only has about 28,000 token pairs.
Swedish has the largest test set with more than 33,000 token pairs. 
The unchanged rate also differs a lot. There are 84.4\% and 75.8\% historical 
spellings that are identical to their modern spellings in German and English, respectively. 
However, the unchanged rate is only 17.1\% in Hungarian. In addition, Hungarian has the largest token vocabulary 
and character vocabulary\footnote{The character vocabulary includes both alphabetic and non-alphabetic characters.}. 
The longest token in Icelandic is only 16, but the longest 
token in Swedish is 36. The average token length of Hungarian is 5.91, which is the 
longest in all five languages. This is because Hungarian is an agglutinative language.

\subsection{Experimental Settings}
Different architectures are hard to compare fairly because many factors 
affect performance. We aim to create a level playing field for the comparison 
by training with the same toolkit, Marian \cite{mariannmt2018}\footnote{We also trained Transformer models with \textit{Tensor2tensor} \url{https://github.com/tensorflow/tensor2tensor}, which achieved better performance in German, Icelandic, and Swedish.}. 
Since there is no implementation of models without attention in Marian, 
we modify the decoder part to enable Marian to train models without attention.\footnote{The modification, the NMT model settings, and the code are available in \url{https://github.com/tanggongbo/normalization-NMT}} 
We assume that the case of letters is useful for predicting the modern spellings. 
Thus, the letters in the training set and the tuning set are case-sensitive. 
The historical spellings in the test set which are the inputs of the model are also 
case-sensitive. However, to keep consistency with the baseline, we lowercase all the 
predicted modern spellings during evaluation.

For character-level models, all the characters are added into the vocabulary, 
even if they only appear once.
For subword-level models, we utilize the BPE method in \newcite{sennrich16sub} 
to generate subword units. 
We try different BPE vocabulary sizes, varying between 100, 200, 300, 500, 1,000 and 5,000.

The vanilla RNN chooses the ``tanh'' RNN cell. 
We enable ``mini-bach-fit'' which automatically choose the mini-batch size 
for the given ``workspace'' size, and the ``workspace'' is set to 7500. 
We use \textit{Adam} \cite{Kingma2014AdamAM} as the optimizer. 
The learning rate is set to 0.0003, but we set the warmup steps to 16,000, 
which means that the learning rate increases linearly before 16,000 steps. 
A model checkpoint is saved every 500 updates. 
The evaluation metrics on the development set are cross-entropy and perplexity. 
We set the early stopping patience to 8 checkpoints. 
All the neural networks have 6 layers. 
The size of embeddings is 512. 
We tie the target embeddings and the output embeddings in the output layer. 
We use the checkpoint that achieves the best perplexity to generate the normalizations. 
We set the beam size to 5 during decoding.

\section{Results}
The baseline from \newcite{pettersson2014evaluation} has very high word accuracy 
and low CER scores in all five languages. 
The results in the baseline are obtained using character-level SMT models except for 
Icelandic, where the combination of a Levenshtein-based method and a dictionary-based 
method achieved the best results.
We use word accuracy and CER to evaluate the predictions.
For the historical spelling normalization task, word accuracy is a very important 
evaluation metric. Moreover, word accuracy is the only evaluation metric in 
\newcite{bollmann2016improve} and \newcite{bollmann2017learning}. 
However, CER is a good supplement to word accuracy. It is more 
fine-grained and evaluates the character-level normalizations. 
In our experiments, we use Levenshtein distance to compute CER. 
Table~\ref{table-result} gives the detailed results of different models in five languages. 

\begin{table*}[ht]
\begin{center}
\begin{tabular}{|c|c|c||c|c||c|c||c|c||c|c|}
\hline \ & \multicolumn{2}{c||}{\bf English} & \multicolumn{2}{c||}{\bf German} & \multicolumn{2}{c||}{\bf Hungarian} & \multicolumn{2}{c||}{\bf Icelandic} & \multicolumn{2}{c|}{\bf Swedish}\\ 
\cline{2-11} \ & Acc & CER &  Acc & CER &  Acc & CER &  Acc & CER &  Acc & CER  \\ 
\hline \ Baseline & 94.3&0.07 &96.6 & 0.04&80.1 &0.21 & 84.6&0.19 &\bf 92.9 &0.07  \\ 
\hline
\ NoAtt-RNN & 94.73& 0.02& 94.89& 0.02&90.99 & 0.03&86.73 & 0.05&91.44 & 0.03\\ 
 \ NoAtt-GRU &94.79 & 0.02&94.85 & 0.02&91.03 & 0.03&86.98 & 0.05&91.34 & 0.03\\ 
 \ NoAtt-LSTM & 94.61& 0.02&95.78 & 0.02&90.91 & 0.03&86.61 & 0.05&91.29 & 0.03\\ 
\hline
  Att-RNN & 94.69& 0.02 & 94.23&0.02& 91.69& 0.02&\bf 87.59& 0.04&91.56 & 0.03\\ 
 \ Att-GRU & 94.80& 0.02 &94.83 &0.02&91.68 & 0.02&87.17 & 0.05&91.68 & 0.03\\ 
 \ Att-LSTM & 94.85& 0.02&96.00 & 0.02&91.57 & 0.03&86.83 & 0.05&91.72 & 0.03\\ 
\hline
Transformer &\bf 95.16&0.02 & 95.22& 0.02&\bf 92.14 & 0.02&86.45 & 0.05&88.99 & 0.05\\ 
\hline
\ BPE-Soft & 95.02&0.02 &\bf 96.64& 0.01& 91.96&0.03 & 87.19& 0.03&91.21 & 0.03\\ 
\hline
\end{tabular}
\caption{\label{table-result} Evaluation results in word accuracy (Acc, \%) and CER. 
The best results in each language have background color. 
Many identical values in CER are different, but the difference is irrelevant in Chi-square test.
}\end{center}
\end{table*}

\subsection{Word Accuracy}
Table~\ref{table-result} shows that NMT models outperform SMT models in four out of 
five languages, except for Swedish, when we use word accuracy as the evaluation 
metric. Compared to the other four languages, we get a huge absolute improvement of 
12.04\% in Hungarian, improving the word accuracy from 80.1\% to 92.14\%. We get 0.04\%, 
0.86\%, and 2.99\% absolute improvement in German, English, and Icelandic, 
respectively. Our best NMT result in Swedish is still a little lower than 
the baseline in word accuracy. 
We attribute the reason to the dataset size, because Swedish has the smallest training set.

We divide the incorrectly normalized spellings into three groups by checking the normalizations of the test set automatically:
\begin{enumerate}[noitemsep]
\item \textit{Change}: modern spelling is identical to historical spelling, 
but the model normalized the historical spelling to another spelling. 
\item \textit{Copy}: modern spelling is different from historical spelling, 
but the model copied the historical spelling as the normalization. 
\item \textit{Other}: other types of error. 
\end{enumerate}

\begin{table}[htbp]
\begin{center}
\begin{tabular}{|c|c|c|c|c|c|}
\hline \  &\bf English &\bf German &\bf Hungarian&\bf Icelandic &\bf Swedish\\ 
\hline \bf Change  & 22.3&28.5&6.1&33.8&25.0\\
\hline \bf Copy &  22.7 &41.7 &6.1&20.8&23.6\\
\hline \bf Other &55 &29.8&87.8&45.4&51.4\\
\hline
\end{tabular}
\caption{\label{table-error-ditribution} Error distributions (\%).}
\end{center}
\end{table}

\noindent 
Table~\ref{table-error-ditribution} gives us the error distributions of the 
best model in each language. 
The \textit{Change} and \textit{Copy} errors only account for 12.2\% in 
Hungarian which is reasonable, because the changed rate in Hungarian is only 17.1\%.
The other four languages still have a lot of \textit{Change} and \textit{Copy} errors.
This finding reveals that it is a little bit difficult for the NMT model trained on the 
data that mixed with changed and unchanged spellings to normalize unchanged spellings. 
However, there are only very few unchanged translations in the MT task. 

Therefore, we explore a hybrid method, combining the NMT-based method 
and the dictionary-based method. More specifically, we first extract a list of 
unchanged spellings from the training set. During the evaluation, 
if a word is in this list, we simply copy it as its normalization. 
If it is not in the list, we feed it to the NMT models. 
The results in Table \ref{table-res-hybrid} show that this hybrid method improves 
the accuracy further. In particular, the improvements on Icelandic are around 5\%. 

\begin{table*}[ht]
\begin{center}
\begin{tabular}{|c|c|c||c|c||c|c||c|c||c|c|}
\hline \ & \multicolumn{2}{c||}{\bf English} & \multicolumn{2}{c||}{\bf German} & \multicolumn{2}{c||}{\bf Hungarian} & \multicolumn{2}{c||}{\bf Icelandic} & \multicolumn{2}{c|}{\bf Swedish}\\ 
\cline{2-11} \ & Acc & $\bigtriangleup$ &  Acc & $\bigtriangleup$ &  Acc & $\bigtriangleup$ &  Acc & $\bigtriangleup$ &  Acc & $\bigtriangleup$  \\ 
\hline
\ NoAtt-RNN & 95.92& 1.19& 95.78& 0.90&91.81 & 0.82&91.92 & 5.18&91.83 &0.39  \\ 
 \ NoAtt-GRU &95.93 & 1.14&95.44 & 0.60&91.87 & 0.84&91.70 & 4.71&91.73 &0.39  \\ 
 \ NoAtt-LSTM & 95.81& 1.20&96.42 & 0.64&91.75 & 0.83&91.78 & 5.17&91.69 &0.41  \\ 
\hline
  Att-RNN & 95.90& 1.21& 94.93& 0.70& 92.47& 0.77& 92.54&4.95 &91.94 &0.38  \\ 
 \ Att-GRU & 95.99&1.19 &95.48 & 0.66&92.49 &0.82 &92.25 &5.08 &92.04 &0.36  \\ 
 \ Att-LSTM & 96.02&1.17 & 96.44& 0.44&92.36 & 0.78&91.76 &4.93 &92.08 &0.36  \\ 
\hline
Transformer & 96.33& 1.17& 95.70&0.48 &92.94 & 0.80&91.60 &5.15 &89.48 & 0.49 \\ 
\hline
\ BPE-Soft & 96.19& 1.18& 96.96& 0.32& 92.74& 0.78& 92.14& 4.95&91.56 &0.35 \\ 
\hline
\end{tabular}
\caption{\label{table-res-hybrid} The results of combining the NMT-based method and the dictionary-based method. "$\bigtriangleup$" denotes the absolute improvement on accuracy (\%) compared to the NMT-based method. 
}\end{center}
\end{table*}

\subsection{CER}
With the CER measure, we calculate the number of correctly normalized characters, without considering the word level.
CER is similar to the BLEU score in MT, and we evaluate at sub-sequence-level 
rather than the overall accuracy. 
When we use CER as the evaluation metric, NMT models get the best results for all 
five languages, even though some models achieve lower accuracy than the baseline. 
This result is different from the result of \newcite{korchagina2017normalizing}. 
In her paper, if the SMT models are better than the NMT models in word accuracy, 
these SMT models are better than the NMT models in CER as well.
We assume that this may be due to different neural network architectures: 
they use CNNs while we use RNNs and self-attention networks.

\begin{table}[htbp]
\begin{center}
\begin{tabular}{|c|c|c|c|c|c|}
\hline \  &\bf English &\bf German &\bf Hungarian&\bf Icelandic &\bf Swedish\\ 
\hline \bf Changed  & 1.45&1.07&2.58&1.41&1.32\\
\hline \bf Incorrect &  1.81 &1.64 &1.78&1.64&1.54\\
\hline
\end{tabular}
\caption{\label{table-editdist} The average edit distance of the changed spellings in 
test set and the average edit distance of the incorrectly normalized changed spellings.}
\end{center}
\end{table}

\noindent Table \ref{table-editdist} shows the edit distance of spellings. 
For the incorrectly normalized changed spellings, the average edit distance 
is smaller than 2. In other words, we just need less than two edits 
to translate an incorrectly normalized spelling into the correct one. 
In the incorrect normalizations, Swedish has the shortest average edit distance 1.54, 
and English has the longest average edit distance 1.81.

Intuitively, if a spelling has smaller edit distance, it is easier for the model 
to normalize this spelling correctly. That is to say, the average edit distance of 
incorrectly normalized spellings will be larger compared to the average edit distance 
before normalization. 
However,  Hungarian is the exception in Table ~\ref{table-editdist}, which indicates 
that spellings with longer edit distance are more likely to 
be normalized close to modern spellings in Hungarian.
For example, the edit distance between ``m\=odanac'' and ``mond\'ak'' is 6, yet
the model can normalize it correctly. 
Although the model normalized ``m\.egb\`et\`e\'geitn\d{c}'' into 
``megbeteg\'iteni\'uk'', which is not identical to the modern spelling ``megbeteg\'iten\'ek'', the edit distance nevertheless decreased from 9 to 2.
We hypothesize that this could be due to the fact that Hungarian belongs to a different language family 
than the other four languages.

\begin{table}[htbp]
\begin{center}
\begin{tabular}{|c|c|c|c|c|c|}  
\hline  \ &\bf English&\bf German&\bf Hungarian&\bf Icelandic&\bf Swedish \\
\hline Historical &alys&julius&v\`et\=e &uopn & sielffuer\\
Normalized & alis & jiues & vetem &opnu & sj\"alver \\
Modern &  alice& julius & vet\'em & vopn & sj\"alv \\ 
\hline Historical &wett&coh{\ae}rentz& haila& sier&herrskaper \\
Normalized &wit& cohaerenz& hajola&sjer & herrskaper\\
Modern &know & koh\"arenz &hajla &s\'er & herrskapen \\ 
\hline
\end{tabular}
\caption{\label{table-examples} Some incorrectly normalized examples from the development set.}
\end{center}
\end{table}

Table \ref{table-examples} gives some incorrectly normalized examples 
from the development set. 
Most of the edit distances of spellings are longer than 1. 
In addition to \textit{Change} and \textit{Copy} errors, 
some historical spellings are quite different from their modern spelling, such as 
``wett'' in English. For the historical word ``wett'', 
it is extinct, people just mapped a semantic related modern word to it. 
``know'' has no relations with ``wett'' in spelling and pronunciation.
Characters with different accents also cause mistakes easily. 
For example, ``vet\'em'' in Hungarian and ``s\'er'' in Icelandic.

\subsection{NMT versus SMT}
In the conventional MT tasks, NMT models usually outperform SMT models. The first 
reason is that the dense embeddings in NMT are powerful representations. 
The second reason is that NMT models usually consider a larger context 
compared to SMT models. This is the same in historical spelling normalization.
In our experiments, the most obvious example is Hungarian. 
The absolute improvement is 12.04\% in word accuracy. 
Compared to other languages, Hungarian has the largest token and 
character vocabularies and the highest changed rate. It also has the longest average 
token length. 
Thus, NMT models can represent these larger vocabularies better than SMT models. 
NMT models are also better at capturing the context information when generating 
the normalization. For example, the NMT models can normalize a 14-character spelling 
``aldozatt'oknak'' into ``\'aldoza\textbf{tuk}nak'' correctly, while the SMT models 
normalize it into ``\'aldoza\textbf{tok}nak''. In the training set, `tok' is much 
more frequent than `tuk'. Since SMT models are more focused on a local context, 
the SMT models choose `tok' rather than `tuk'.

However, in terms of accuracy, it is still hard for NMT models to exceed SMT models 
in Swedish. We also find that the performance of NMT models is quite close to the 
baseline in German which has the second smallest training dataset. 
We hypothesize that the size of training data is crucial for NMT models to exceed SMT models. 

As there is much more test data in Swedish compared to other languages, 
we test our hypothesis by moving some token pairs from test sets to training sets 
and development sets. More specifically, we create two new datasets, in which 27,000 
and 30,000 token pairs are moved from the beginning of the test set to the training set and the 
development set, respectively. 
Both the datasets and results are described in Table \ref{table-se}. 

\begin{table}[htbp]
\begin{center}
\begin{tabular}{|c|c|c|c|c|c|c|}
\hline Training &Development &Test&Att-RNN &Att-GRU&Att-LSTM&Transformer\\
\hline 28,327& 2,590 &33,544 & 91.56&91.68 &91.72 & 88.99\\
\hline 51,237& 6,590 & \phantom{0}6,544& 95.45& 94.79&94.97 &95.18 \\
\hline 57,637& 3,190 & \phantom{0}3,544& 96.02& 95.77& 95.65& 95.77\\
\hline
\end{tabular}
\caption{\label{table-se} The accuracy of different models in Swedish with different dataset settings.}
\end{center}
\end{table}

\noindent Table \ref{table-se} shows that the NMT models achieve much higher accuracy 
with more training data. This result indicates that the performance of NMT models is 
highly related to the size of training set. 

\subsection{Different NMT Models}

Hypothesis 1 is that the performance gap between vanilla RNNs and 
GRUs/LSTMs will not be huge. The results in Table \ref{table-result} reveal that 
the vanilla RNNs are competitive to the GRUs/LSTMs in this task. 
\textit{Att-RNN} even performs better than \textit{Att-GRU/LSTM} in Icelandic. 
However, \textit{Att-RNN} is clearly worse than \textit{Att-LSTM} in German. 
These results support our Hypothesis 1 well.

Hypothesis 2 states that NMT models with and without attention will not differ 
a lot. The models with attention are slightly better than models without attention 
in our experiments, which is in line with the results in \newcite{bollmann2017learning}. 
However, the gap is quite small. Thus, it fits our Hypothesis 2. 

Hypothesis 3 is that Transformer models are better than 
soft-attention-based models. From Table \ref{table-result}, we can see that 
\textit{Transformer}, with self-attention and multi-head attention, achieves higher 
word accuracy in English and Hungarian compared to soft-attention-based models. 
It is interesting that English and Hungarian have much more training data compared 
to the other three languages. This result reveals that Transformer models need 
more data to exceed RNN-based models. 

Hypothesis 4, finally, states that subword-level models are better than character-level models. 
Our experimental results of \textit{BPE-Soft} models in four languages (except Swedish) 
show that subword-level models are superior to character-level models when the BPE 
vocabulary is small. In subword-level models, 
the vocabulary includes all the characters and learned subword units. We try 
different BPE vocabulary sizes. All the subword-level models are trained on LSTMs. 
Table~\ref{table-bpe} gives the detailed results with different BPE sizes. 

Many historical spellings only have several instances in the training set. 
The NMT model cannot translate the token well at the token level. Moreover, 
there is also a data sparsity problem for the subwords when we set a larger BPE vocabulary. 
We assume that BPE maybe cannot learn rare subword units very well, 
because of the data sparsity. 
That is why subword-level models perform better in the conventional MT tasks, 
which have a much larger training set.
We find that the subword-level models perform worse than character-level models when 
the BPE vocabulary is larger than 300 in all five languages. 

We further train Transformer models at subword-level which are called \textit{
BPE-Transformer} in Table \ref{table-bpe}. In German, Icelandic, and Swedish 
where the data size is small, the subword-level models surpass character-level models. 
However, the subword-level models in English and Hungarian are clearly not as well as character-level models. 

\begin{table}[htbp]
\begin{center}
\begin{tabular}{|c|c|c|c|c|c|c|}
\hline \bf Model &\bf BPE-size &\bf English &\bf German&\bf Hungarian &\bf Icelandic&\bf Swedish\\
\hline \multirow{7}{*}{BPE-Soft}&0& 94.85 & 96.00 & 91.57 & 86.83 & \bf 91.72 \\
 &100& \bf 95.02 & \bf 96.64 &  91.87 &\bf  87.19 & 91.21\\
&200& 94.91 & 96.28 & 91.81 & 86.89 & 91.13 \\
&300& 94.69 & 96.50 & \bf91.96 & 86.76 & 90.84 \\
&500& 94.54 & 96.42 & 91.52 & 86.51 & 90.57 \\
&1,000& 94.52 & 96.18 & 91.44 & 86.29 & 89.67 \\
& 5,000& 93.71 & 95.06 & 89.43 & 84.87 & 85.47 \\
\hline \multirow{4}{*}{BPE-Transformer}&0& \bf 95.16 & 95.22& \bf 92.14 &86.45&88.99\\ 
&100 &94.21 &95.66 &90.14 & \bf 86.64 &90.07 \\
&200 &94.38 &96.08 &90.71 &86.62 & \bf 90.17 \\
&300 &94.26& \bf 96.10 &90.87 &86.33 &89.76 \\
\hline
\end{tabular}
\caption{\label{table-bpe} Accuracy (\%) with different BPE vocabulary sizes. ``0'' represents the character-level models.}
\end{center}
\end{table}

\noindent Historical languages which have little training data are considered as 
low-resource languages, especially the German, the Icelandic, and the Swedish 
in this paper. Hence the result of Hypothesis 4 can be interpreted as that 
subword-level models with a small subword vocabulary can further improve the 
performance compared to character-level models in low-resource languages.

\section{Conclusions and Future Work}
In this paper, we explore different NMT models for the historical spelling normalization 
task in five languages, English, German, Hungarian, Icelandic, and Swedish. 
We propose four hypotheses on NMT models, which are the general questions to ask 
when applying NMT models to the historical spelling normalization task. 

We find that the performance gap between vanilla RNNs and GRUs/LSTMs is very 
small, vanilla RNNs are even competitive to GRUs/LSTMs in Hungarian and Icelandic. 
We demonstrate that the gap between NMT models with or without attention is also slight. 
We show that the subword-level models with a small subword vocabulary are better 
than character-level models for low-resource languages. However, subword-level models 
with a larger vocabulary suffer from data sparsity. 

When we use word accuracy as the evaluation metric, NMT models can get better results 
for four languages compared to SMT models. However, all the NMT models perform 
better than SMT models for all five languages when we use CER as the evaluation metric. 
In addition, the size of the training set is crucial to NMT models. Particularly, 
Transformer models are superior to RNN-based models only when provided with more training data. 
These findings could contribute to the development of general NMT systems, 
especially for low-resource languages. 
Since NMT models are more likely to generate incorrect normalizations of unchanged spellings, 
we propose a hybrid method using both NMT-based methods and dictionary-based method 
which improves the performance further. 

In the future, we could 1) explore some hard-attention-based models, 2) introduce phoneme knowledge into NMT models, and 3) use sentence pairs for spelling normalization.
Compared to soft attention, hard attention \cite{xu2015show} only pays attention to one or several specified source word annotations. \newcite{aharoni2017morphological} employ hard monotonic attention for a morphological inflection generation task. 
The variation between historical spelling and modern spelling is usually monotonic, which is similar to morphological inflection. 
Thus, hard attention should work well in historical spelling normalization as well. 

Many words have changed their spellings, but they keep the same pronunciation. Thus, \newcite{bollmann2017learning} use an additional grapheme-to-phoneme dictionary in a multi-task learning setting. We can add the phonetic dictionaries as additional training data to improve the performance. 

In addition to token-pair-based normalization, \newcite{ljubesic16normalising} use segment pairs with context information to do spelling normalization. NMT models are powerful in using context information. Thus, training the NMT models on sentence pairs is likely to improve the spelling normalization task further, which introduces more context information.

\section*{Acknowledgments}
We thank all the anonymous reviews who give a lot of valuable and insightful comments. 
We acknowledge the computational resources provided by CSC in Helsinki and Sigma2 in Oslo
through NeIC-NLPL (www.nlpl.eu). We also thank the machine translation group at the University of Edinburgh for providing  computational resources. 
Gongbo Tang is funded by Chinese Scholarship Council (NO. 201607110016). 

\bibliographystyle{acl}
\bibliography{coling2018}

\begin{thebibliography}{}

\bibitem[\protect\citename{Aharoni and Goldberg}2017]{aharoni2017morphological}
Roee Aharoni and Yoav Goldberg.
\newblock 2017.
\newblock Morphological inflection generation with hard monotonic attention.
\newblock In {\em Proceedings of the 55th Annual Meeting of the Association for
  Computational Linguistics (Volume 1: Long Papers)}, pages 2004--2015,
  Vancouver, Canada. Association for Computational Linguistics.

\bibitem[\protect\citename{Bahdanau \bgroup et al.\egroup
  }2015]{bahdanau15joint}
Dzmitry Bahdanau, Kyunghyun Cho, and Yoshua Bengio.
\newblock 2015.
\newblock Neural machine translation by jointly learning to align and
  translate.
\newblock In {\em Proceedings of the 3rd International Conference on Learning
  Representations}, San Diego, California, USA.

\bibitem[\protect\citename{Bollmann and S{\o}gaard}2016]{bollmann2016improve}
Marcel Bollmann and Anders S{\o}gaard.
\newblock 2016.
\newblock Improving historical spelling normalization with bi-directional lstms
  and multi-task learning.
\newblock In {\em Proceedings of COLING 2016, the 26th International Conference
  on Computational Linguistics: Technical Papers}, pages 131--139, Osaka,
  Japan. The COLING 2016 Organizing Committee.

\bibitem[\protect\citename{Bollmann \bgroup et al.\egroup
  }2011]{bollmann2011rule}
Marcel Bollmann, Florian Petran, and Stefanie Dipper.
\newblock 2011.
\newblock Rule-based normalization of historical texts.
\newblock In {\em Proceedings of the Workshop on Language Technologies for
  Digital Humanities and Cultural Heritage}, pages 34--42, Hissar, Bulgaria.
  Association for Computational Linguistics.

\bibitem[\protect\citename{Bollmann \bgroup et al.\egroup
  }2017]{bollmann2017learning}
Marcel Bollmann, Joachim Bingel, and Anders S{\o}gaard.
\newblock 2017.
\newblock Learning attention for historical text normalization by learning to
  pronounce.
\newblock In {\em Proceedings of the 55th Annual Meeting of the Association for
  Computational Linguistics (Volume 1: Long Papers)}, pages 332--344,
  Vancouver, Canada. Association for Computational Linguistics.

\bibitem[\protect\citename{Bollmann}2013]{bollmann2013pos}
Marcel Bollmann.
\newblock 2013.
\newblock Pos tagging for historical texts with sparse training data.
\newblock In {\em Proceedings of the 7th Linguistic Annotation Workshop and
  Interoperability with Discourse}, pages 11--18, Sofia, Bulgaria. Association
  for Computational Linguistics.

\bibitem[\protect\citename{Britz \bgroup et al.\egroup }2017]{britz2017massive}
Denny Britz, Anna Goldie, Thang Luong, and Quoc Le.
\newblock 2017.
\newblock Massive exploration of neural machine translation architectures.
\newblock {\em arXiv preprint arXiv:1703.03906}.

\bibitem[\protect\citename{Cho \bgroup et al.\egroup }2014]{cho2014learning}
Kyunghyun Cho, Bart van Merrienboer, Caglar Gulcehre, Dzmitry Bahdanau, Fethi
  Bougares, Holger Schwenk, and Yoshua Bengio.
\newblock 2014.
\newblock Learning phrase representations using {RNN} encoder--decoder for
  statistical machine translation.
\newblock In {\em Proceedings of the 2014 Conference on Empirical Methods in
  Natural Language Processing}, pages 1724--1734, Doha, Qatar. Association for
  Computational Linguistics.

\bibitem[\protect\citename{Chung \bgroup et al.\egroup
  }2016]{chung2016character}
Junyoung Chung, Kyunghyun Cho, and Yoshua Bengio.
\newblock 2016.
\newblock A character-level decoder without explicit segmentation for neural
  machine translation.
\newblock In {\em Proceedings of the 54th Annual Meeting of the Association for
  Computational Linguistics (Volume 1: Long Papers)}, pages 1693--1703, Berlin,
  Germany. Association for Computational Linguistics.

\bibitem[\protect\citename{Costa-juss\`{a} and
  Fonollosa}2016]{Costa2016Character}
Marta~R. Costa-juss\`{a} and Jos\'{e} A.~R. Fonollosa.
\newblock 2016.
\newblock Character-based neural machine translation.
\newblock In {\em Proceedings of the 54th Annual Meeting of the Association for
  Computational Linguistics (Volume 2: Short Papers)}, pages 357--361, Berlin,
  Germany. Association for Computational Linguistics.

\bibitem[\protect\citename{Fiebranz \bgroup et al.\egroup
  }2011]{fiebranz2011making}
Rosemarie Fiebranz, Erik Lindberg, Jonas Lindstr{\"o}m, and Maria {\AA}gren.
\newblock 2011.
\newblock Making verbs count: the research project ‘gender and work’and its
  methodology.
\newblock {\em Scandinavian Economic History Review}, 59(3):273--293.

\bibitem[\protect\citename{Gage}1994]{gage1994new}
Philip Gage.
\newblock 1994.
\newblock A new algorithm for data compression.
\newblock {\em The C Users Journal}, 12(2):23--38.

\bibitem[\protect\citename{Hochreiter and Schmidhuber}1997]{hochreiter1997long}
Sepp Hochreiter and J{\"u}rgen Schmidhuber.
\newblock 1997.
\newblock Long short-term memory.
\newblock {\em Neural computation}, 9(8):1735--1780.

\bibitem[\protect\citename{Junczys-Dowmunt \bgroup et al.\egroup
  }2018]{mariannmt2018}
Marcin Junczys-Dowmunt, Roman Grundkiewicz, Tomasz Dwojak, Hieu Hoang, Kenneth
  Heafield, Tom Neckermann, Frank Seide, Ulrich Germann, Alham Fikri~Aji,
  Nikolay Bogoychev, Andr\'{e} F.~T. Martins, and Alexandra Birch.
\newblock 2018.
\newblock Marian: Fast neural machine translation in {C++}.
\newblock {\em arXiv preprint arXiv:1804.00344}.

\bibitem[\protect\citename{Kalchbrenner and Blunsom}2013]{kal2013recurrent}
Nal Kalchbrenner and Phil Blunsom.
\newblock 2013.
\newblock Recurrent continuous translation models.
\newblock In {\em Proceedings of the 2013 Conference on Empirical Methods in
  Natural Language Processing}, pages 1700--1709, Seattle, Washington, USA.
  Association for Computational Linguistics.

\bibitem[\protect\citename{Kingma and Ba}2015]{Kingma2014AdamAM}
Diederik~P. Kingma and Jimmy Ba.
\newblock 2015.
\newblock Adam: A method for stochastic optimization.
\newblock In {\em Proceedings of the 3rd International Conference on Learning
  Representations}, San Diego, California, USA.

\bibitem[\protect\citename{Koehn \bgroup et al.\egroup
  }2003]{koehn2003statistical}
Philipp Koehn, Franz~Josef Och, and Daniel Marcu.
\newblock 2003.
\newblock Statistical phrase-based translation.
\newblock In {\em Proceedings of the 2003 Conference of the North American
  Chapter of the Association for Computational Linguistics on Human Language
  Technology-Volume 1}, pages 48--54. Association for Computational
  Linguistics.

\bibitem[\protect\citename{Korchagina}2017]{korchagina2017normalizing}
Natalia Korchagina.
\newblock 2017.
\newblock Normalizing medieval german texts: from rules to deep learning.
\newblock In {\em Proceedings of the NoDaLiDa 2017 Workshop on Processing
  Historical Language}, number 133, pages 12--17, Gothenburg, Sweden.
  Link{\"o}ping University Electronic Press.

\bibitem[\protect\citename{Ling \bgroup et al.\egroup }2015]{ling15char}
Wang Ling, Isabel Trancoso, Chris Dyer, and Alan~W Black.
\newblock 2015.
\newblock Character-based neural machine translation.
\newblock {\em arXiv preprint arXiv:1511.04586}.

\bibitem[\protect\citename{Ljube\v{s}i\'{c} \bgroup et al.\egroup
  }2016]{ljubesic16normalising}
Nikola Ljube\v{s}i\'{c}, Katja Zupan, Darja Fi{\v s}er, and Toma{\v z} Erjavec.
\newblock 2016.
\newblock {Normalising Slovene data: historical texts vs. user-generated
  content}.
\newblock In {\em Proceedings of the 13th Conference on Natural Language
  Processing}, pages 146--155, Varanasi, India. Association for Computational
  Linguistics.

\bibitem[\protect\citename{Markus}1999]{markus1999manual}
Manfred Markus.
\newblock 1999.
\newblock {\em Manual of ICAMET (Innsbruck Computer Archive of Machine-Readable
  English Texts)}.
\newblock Leopold-Franzens-Universitat Innsbruck.

\bibitem[\protect\citename{Papineni \bgroup et al.\egroup
  }2002]{papineni2002bleu}
Kishore Papineni, Salim Roukos, Todd Ward, and Wei-Jing Zhu.
\newblock 2002.
\newblock Bleu: a method for automatic evaluation of machine translation.
\newblock In {\em Proceedings of 40th Annual Meeting of the Association for
  Computational Linguistics}, pages 311--318, Philadelphia, Pennsylvania, USA.
  Association for Computational Linguistics.

\bibitem[\protect\citename{Pettersson \bgroup et al.\egroup
  }2013a]{pettersson2013normalisation}
Eva Pettersson, Be{\'a}ta Megyesi, and Joakim Nivre.
\newblock 2013a.
\newblock Normalisation of historical text using context-sensitive weighted
  levenshtein distance and compound splitting.
\newblock In {\em Proceedings of the 19th Nordic Conference of Computational
  Linguistics}, pages 163--179, Oslo, Norway. Association for Computational
  Linguistics.

\bibitem[\protect\citename{Pettersson \bgroup et al.\egroup
  }2013b]{pettersson2013smt}
Eva Pettersson, Be{\'a}ta Megyesi, and J{\"o}rg Tiedemann.
\newblock 2013b.
\newblock An {SMT} approach to automatic annotation of historical text.
\newblock In {\em Proceedings of the workshop on computational historical
  linguistics at NODALIDA 2013}, pages 54--69, Oslo, Norway. Association for
  Computational Linguistics.

\bibitem[\protect\citename{Pettersson \bgroup et al.\egroup
  }2014]{pettersson2014evaluation}
Eva Pettersson, Be\'{a}ta Megyesi, and Joakim Nivre.
\newblock 2014.
\newblock A multilingual evaluation of three spelling normalisation methods for
  historical text.
\newblock In {\em Proceedings of the 8th Workshop on Language Technology for
  Cultural Heritage, Social Sciences, and Humanities (LaTeCH)}, pages 32--41,
  Gothenburg, Sweden. Association for Computational Linguistics.

\bibitem[\protect\citename{Rayson \bgroup et al.\egroup }2005]{rayson2005vard}
Paul Rayson, Dawn Archer, and Nicholas Smith.
\newblock 2005.
\newblock Vard versus word: A comparison of the {UCREL} variant detector and
  modern spell checkers on english historical corpora.
\newblock In {\em Proceedings of the Corpus Linguistics 2005}, Birmingham, UK.

\bibitem[\protect\citename{R\"{o}gnvaldsson \bgroup et al.\egroup
  }2012]{rognvaldsson2012icelandic}
Eir{\'i}kur R\"{o}gnvaldsson, Anton~Karl Ingason, Einar~Freyr Sigur{\dh}sson,
  and Joel Wallenberg.
\newblock 2012.
\newblock The icelandic parsed historical corpus (icepahc).
\newblock In {\em Proceedings of the 8th International Conference on Language
  Resources and Evaluations}, pages 1977--1984, Istanbul, Turkey, May. European
  Language Resources Association.

\bibitem[\protect\citename{S{\'a}nchez-Mart{\'\i}nez \bgroup et al.\egroup
  }2013]{sanchez2013open}
Felipe S{\'a}nchez-Mart{\'\i}nez, Isabel Mart{\'\i}nez-Sempere, Xavier
  Ivars-Ribes, and Rafael~C Carrasco.
\newblock 2013.
\newblock An open diachronic corpus of historical {S}panish: annotation
  criteria and automatic modernisation of spelling.
\newblock {\em arXiv preprint arXiv:1306.3692}.

\bibitem[\protect\citename{Scheible \bgroup et al.\egroup
  }2011]{scheible2011germanc}
Silke Scheible, Richard~J. Whitt, Martin Durrell, and Paul Bennett.
\newblock 2011.
\newblock A gold standard corpus of early modern german.
\newblock In {\em Proceedings of the 5th Linguistic Annotation Workshop}, pages
  124--128, Portland, Oregon, USA, June. Association for Computational
  Linguistics.

\bibitem[\protect\citename{Scherrer and Erjavec}2013]{scherrer2013modernizing}
Yves Scherrer and Toma\v{z} Erjavec.
\newblock 2013.
\newblock Modernizing historical slovene words with character-based smt.
\newblock In {\em Proceedings of the 4th Biennial International Workshop on
  Balto-Slavic Natural Language Processing}, pages 58--62, Sofia, Bulgaria.
  Association for Computational Linguistics.

\bibitem[\protect\citename{Schuster and Nakajima}2012]{schuster2012japanese}
Mike Schuster and Kaisuke Nakajima.
\newblock 2012.
\newblock Japanese and korean voice search.
\newblock In {\em 2012 IEEE International Conference on Acoustics, Speech and
  Signal Processing (ICASSP)}, pages 5149--5152, Kyoto, Japan. IEEE.

\bibitem[\protect\citename{Sennrich \bgroup et al.\egroup }2016]{sennrich16sub}
Rico Sennrich, Barry Haddow, and Alexandra Birch.
\newblock 2016.
\newblock Neural machine translation of rare words with subword units.
\newblock In {\em Proceedings of the 54th Annual Meeting of the Association for
  Computational Linguistics (Volume 1: Long Papers)}, pages 1715--1725, Berlin,
  Germany. Association for Computational Linguistics.

\bibitem[\protect\citename{Simon}2014]{simon2014corpus}
Eszter Simon.
\newblock 2014.
\newblock Corpus building from old hungarian codices.
\newblock In {\em The Evolution of Functional Left Peripheries in Hungarian
  Syntax}, pages 224--236. Oxford University Press.

\bibitem[\protect\citename{Sutskever \bgroup et al.\egroup
  }2014]{sutskever2014sequence}
Ilya Sutskever, Oriol Vinyals, and Quoc~V Le.
\newblock 2014.
\newblock Sequence to sequence learning with neural networks.
\newblock In {\em Advances in Neural Information Processing Systems 27}, pages
  3104--3112. Curran Associates, Inc., Montr\'eal, Canada.

\bibitem[\protect\citename{Vaswani \bgroup et al.\egroup
  }2017]{vaswani2017Attention}
Ashish Vaswani, Noam Shazeer, Niki Parmar, Jakob Uszkoreit, Llion Jones,
  Aidan~N Gomez, \L~ukasz Kaiser, and Illia Polosukhin.
\newblock 2017.
\newblock Attention is all you need.
\newblock In {\em Advances in Neural Information Processing Systems 30}, pages
  6000--6010. Curran Associates, Inc.

\bibitem[\protect\citename{Wu \bgroup et al.\egroup }2016]{wu2016google}
Yonghui Wu, Mike Schuster, Zhifeng Chen, Quoc~V Le, Mohammad Norouzi, Wolfgang
  Macherey, Maxim Krikun, Yuan Cao, Qin Gao, Klaus Macherey, et~al.
\newblock 2016.
\newblock Google's neural machine translation system: Bridging the gap between
  human and machine translation.
\newblock {\em arXiv preprint arXiv:1609.08144}.

\bibitem[\protect\citename{Xu \bgroup et al.\egroup }2015]{xu2015show}
Kelvin Xu, Jimmy Ba, Ryan Kiros, Kyunghyun Cho, Aaron Courville, Ruslan
  Salakhudinov, Rich Zemel, and Yoshua Bengio.
\newblock 2015.
\newblock Show, attend and tell: Neural image caption generation with visual
  attention.
\newblock In {\em Proceedings of the International Conference on Machine
  Learning}, pages 2048--2057, Lille, France. PMLR.

\end{thebibliography}

\end{document}